\documentclass[]{fairmeta}

\usepackage{tikz}
\usetikzlibrary{shadings}

\definecolor{sprcolor}{RGB}{28, 84, 166}
\definecolor{springgreen}{RGB}{54, 159, 91}
\definecolor{springpink}{RGB}{226, 93, 151}
\definecolor{springcyan}{RGB}{40, 150, 170}
\definecolor{springblue}{RGB}{34, 124, 194}
\definecolor{springgold}{RGB}{242, 176, 75}

\newcommand{\gaussianball}{\tikz[baseline=-0.62ex, x=1ex, y=1ex]{\fill[springblue, opacity=0.09, rotate around={18:(-0.72,-0.18)}] (-0.72,-0.18) ellipse (0.48 and 0.27); \fill[springgreen, opacity=0.10, rotate around={-22:(-0.42,-0.30)}] (-0.42,-0.30) ellipse (0.52 and 0.28); \fill[springpink, opacity=0.08, rotate around={48:(-0.30,0.04)}] (-0.30,0.04) ellipse (0.34 and 0.20); \draw[springblue!75!black, line width=0.04ex, opacity=0.58, ->] (-0.42,0.06) -- (-0.18,0.23); \draw[springgreen!75!black, line width=0.04ex, opacity=0.58, ->] (0.16,-0.12) -- (0.42,-0.28); \draw[springpink!75!black, line width=0.038ex, opacity=0.52, ->] (0.22,0.24) -- (0.10,0.52); \draw[springgold!75!black, line width=0.038ex, opacity=0.50, ->] (-0.06,0.36) -- (-0.30,0.48); \draw[springblue!60!springgreen!80!black, line width=0.038ex, opacity=0.50, ->] (0.54,0.02) -- (0.80,0.16); \fill[springblue, opacity=0.34, rotate around={18:(-0.42,0.06)}] (-0.42,0.06) ellipse (0.56 and 0.32); \fill[springgreen, opacity=0.34, rotate around={-22:(0.16,-0.12)}] (0.16,-0.12) ellipse (0.62 and 0.34); \fill[springpink, opacity=0.28, rotate around={48:(0.22,0.24)}] (0.22,0.24) ellipse (0.42 and 0.25); \fill[springgold, opacity=0.30, rotate around={-10:(-0.06,0.36)}] (-0.06,0.36) ellipse (0.36 and 0.20); \fill[springblue!70!springgreen, opacity=0.26, rotate around={28:(0.54,0.02)}] (0.54,0.02) ellipse (0.34 and 0.21); \draw[springblue!75!black, line width=0.045ex, opacity=0.70, rotate around={18:(-0.42,0.06)}] (-0.42,0.06) ellipse (0.56 and 0.32); \draw[springgreen!75!black, line width=0.045ex, opacity=0.70, rotate around={-22:(0.16,-0.12)}] (0.16,-0.12) ellipse (0.62 and 0.34); \draw[springpink!75!black, line width=0.045ex, opacity=0.60, rotate around={48:(0.22,0.24)}] (0.22,0.24) ellipse (0.42 and 0.25); \draw[springgold!75!black, line width=0.04ex, opacity=0.55, rotate around={-10:(-0.06,0.36)}] (-0.06,0.36) ellipse (0.36 and 0.20); \draw[springblue!70!springgreen!70!black, line width=0.04ex, opacity=0.55, rotate around={28:(0.54,0.02)}] (0.54,0.02) ellipse (0.34 and 0.21); \fill[white, opacity=0.45] (-0.48,0.17) circle (0.07); \fill[white, opacity=0.38] (0.04,0.03) circle (0.06); \fill[white, opacity=0.34] (0.24,0.30) circle (0.05);}}

\DeclareRobustCommand{\methodname}{%
    GaussianDream%
    \kern0.01em%
    \raisebox{0.12ex}[0pt][0pt]{%
        \scalebox{0.70}{++}%
    }%
}
\newcommand{\method}{\methodname\xspace}

\title{
    \vspace{-1pt}
    \gaussianball\hspace{0.25em}
    \textbf{\methodname}:
    Efficient 3D Gaussian World Modeling
    for Robotic Manipulation
}

\author[2,3,1,*]{Yuqing Jiang}
\author[2,3,1,*]{Zijian Zhang}
\author[4]{Weitao Zhou}
\author[5]{Jiawei Wang}
\author[6]{Junjie He}
\author[7]{Lei Yang}
\author[1]{Haifang Qing}
\author[8]{Si Liu}
\author[9]{Ding Zhao}
\author[10]{Ping Luo}

\author[10,1,\ddagger]{Haibao Yu}

\affiliation[1]{Tuojing Intelligence}
\affiliation[2]{University of Chinese Academy of Sciences}
\affiliation[3]{Institute of Automation, Chinese Academy of Sciences}
\affiliation[4]{Tsinghua University}
\affiliation[5]{University of Science and Technology of China}
\affiliation[6]{The Hong Kong University of Science and Technology (Guangzhou)}
\affiliation[7]{Nanyang Technological University }
\affiliation[8]{Beihang University}
\affiliation[9]{Carnegie Mellon University}

\affiliation[10]{The University of Hong Kong}

\contribution[*]{equal contribution}
\contribution[\ddagger]{corresponding author}

\usepackage{amsmath,amsfonts,bm}
\usepackage{xcolor}

\def\eqref#1{equation~\ref{#1}}

\def\1{\bm{1}}

\DeclareMathAlphabet{\mathsfit}{\encodingdefault}{\sfdefault}{m}{sl}
\SetMathAlphabet{\mathsfit}{bold}{\encodingdefault}{\sfdefault}{bx}{n}

\usepackage{graphicx}
\usepackage[table]{xcolor}
\usepackage{colortbl}

\usepackage{booktabs}
\usepackage{tabularx}
\usepackage{array}
\usepackage{multirow}
\usepackage{diagbox}
\usepackage{hhline}
\usepackage{longtable}
\usepackage{makecell}
\usepackage{siunitx}
\usepackage{adjustbox}
\usepackage{wrapfig}
\usepackage{caption}   
\newcolumntype{C}{>{\centering\arraybackslash}X}

\usepackage{amsmath,amsfonts,amssymb}
\usepackage{bm}
\usepackage{nicefrac}

\usepackage{enumitem}
\setlist[itemize]{leftmargin=*}

\usepackage{caption}
\crefname{figure}{Fig.}{Figs.}
\crefname{table}{Tab.}{Tabs.}

\usepackage{xspace}
\usepackage{calc}
\usepackage{etoolbox}
\usepackage{pifont}
\usepackage{fancyvrb}

\usepackage{titletoc}

\titlecontents{section}
[1.5em] %
{\addvspace{-0.5pt}} %
{\bfseries\contentslabel{2.3em}} %
{\hspace*{-2.3em}\bfseries} %
{\bfseries\titlerule*[.5pc]{.}\contentspage} %
\titlecontents{subsection}
[3.8em] %
{\addvspace{-2.2pt}} %
{\contentslabel{2.3em}}
{\hspace*{-2.3em}}
{\titlerule*[.5pc]{.}\contentspage}

\abstract{
Vision-Language-Action (VLA) policies have advanced language-conditioned
robotic manipulation, yet action-imitation objectives provide only weak
supervision for metric 3D structure and short-horizon physical evolution.
Geometry-enhanced policies mainly improve current-scene grounding, whereas
predictive policies often model future dynamics in RGB or latent spaces and
may incur substantial deployment cost. GaussianDream demonstrates that
training-time current Gaussian reconstruction and future Gaussian prediction
provide effective 3D supervision, but its dense VGGT/TGE-based prefix jointly
carries state, dynamics, and action-conditioning information. We present
\textbf{\methodname}, a compact, policy-native extension that inserts
\textbf{World State Tokens} and \textbf{World Prediction Tokens} directly
into the VLA backbone. A training-only \textbf{World Representation Head}
decodes these tokens into a Current World and coupled Future Prediction over
shared Gaussian primitives, while static--dynamic factorization preserves
persistent structure and focuses residual motion on interaction-relevant
regions. At inference, the head, renderer, auxiliary objectives, and
VGGT/TGE pathway are removed, leaving only 20 world tokens without online
Gaussian decoding or rollout. \method achieves \textbf{98.6\%} on LIBERO
and \textbf{87.8\%} on LIBERO-Plus, with clear gains under Camera and Layout
shifts. Real-robot experiments further improve average success from 29.2\%
to 52.5\% over reproduced $\pi_{0.5}$ while maintaining efficient
closed-loop control.
}
\metadata[Code]{\url{https://github.com/TuojingAI/GaussianDream}}
\metadata[Project Website]{\url{https://tuojingai.github.io/GaussianDream-Series-project-page/}}

\definecolor{lightgray}{rgb}{0.95, 0.95, 0.95}

\definecolor{baselinecolor}{gray}{.9}

\begin{document}
\maketitle 

\section{Introduction}
\label{sec:intro}

General-purpose robotic manipulation requires an embodied agent to interpret
language goals and convert visual observations into precise physical actions.
Recent Vision-Language-Action (VLA) policies have made substantial progress
toward this goal by transferring semantic priors from large-scale
vision-language pretraining to robot control
~\citep{brohan2023rt2,kim2024openvla,black2024pi0,
physicalintelligence2025pi05}. Combined with large-scale robot
demonstrations, these priors improve instruction following, object-level
generalization, and continuous action generation. Nevertheless, successful
manipulation also depends on metric 3D structure, object-relative geometry,
and short-horizon physical evolution, all of which are only indirectly
constrained by action labels.

Despite their strong semantic capabilities, current VLA paradigms still face
three challenges in learning such physical structure. First, scene geometry,
visibility, and contact-relevant spatial relationships are commonly encoded
implicitly in 2D visual features, making policies sensitive to subtle errors
in grasping and placement. Second, robot trajectories contain dense evidence
of object layout, depth, appearance, and motion, whereas behavior cloning
primarily supervises the target action and leaves much of this evidence
unused. Third, current-state understanding and future evolution are often
learned through either separate auxiliary objectives or a single entangled
representation, even though short-horizon manipulation is dominated by
persistent scene structure and comparatively sparse interaction-induced
changes.

Recent approaches address parts of these challenges. Geometry-enhanced VLAs
introduce depth, point clouds, calibrated camera cues, stereo observations,
or pretrained 3D representations to strengthen spatial grounding
~\citep{zhen20243d,sun2025geovla,li2025spatialforcing,
peng2026g3vla,chen2026pointact,song2026vga}. These methods improve current
scene anchoring, but usually provide limited supervision for how the physical
state evolves after interaction. Predictive policies and World-Action Models
instead learn from future observations, trajectories, video generation, or
latent dynamics
~\citep{cen2025worldvla,bi2025motus,kim2026cosmospolicy,
ye2026dreamzero,yuan2026fast}. Although future modeling provides richer
temporal supervision, visually plausible predictions in RGB or latent space
do not necessarily encode metrically consistent physical transitions.
Moreover, retaining large geometry encoders, generative models, or explicit
future rollout can increase the cost of closed-loop control.

\begin{figure*}[t]
    \centering
    \includegraphics[width=0.98\textwidth]{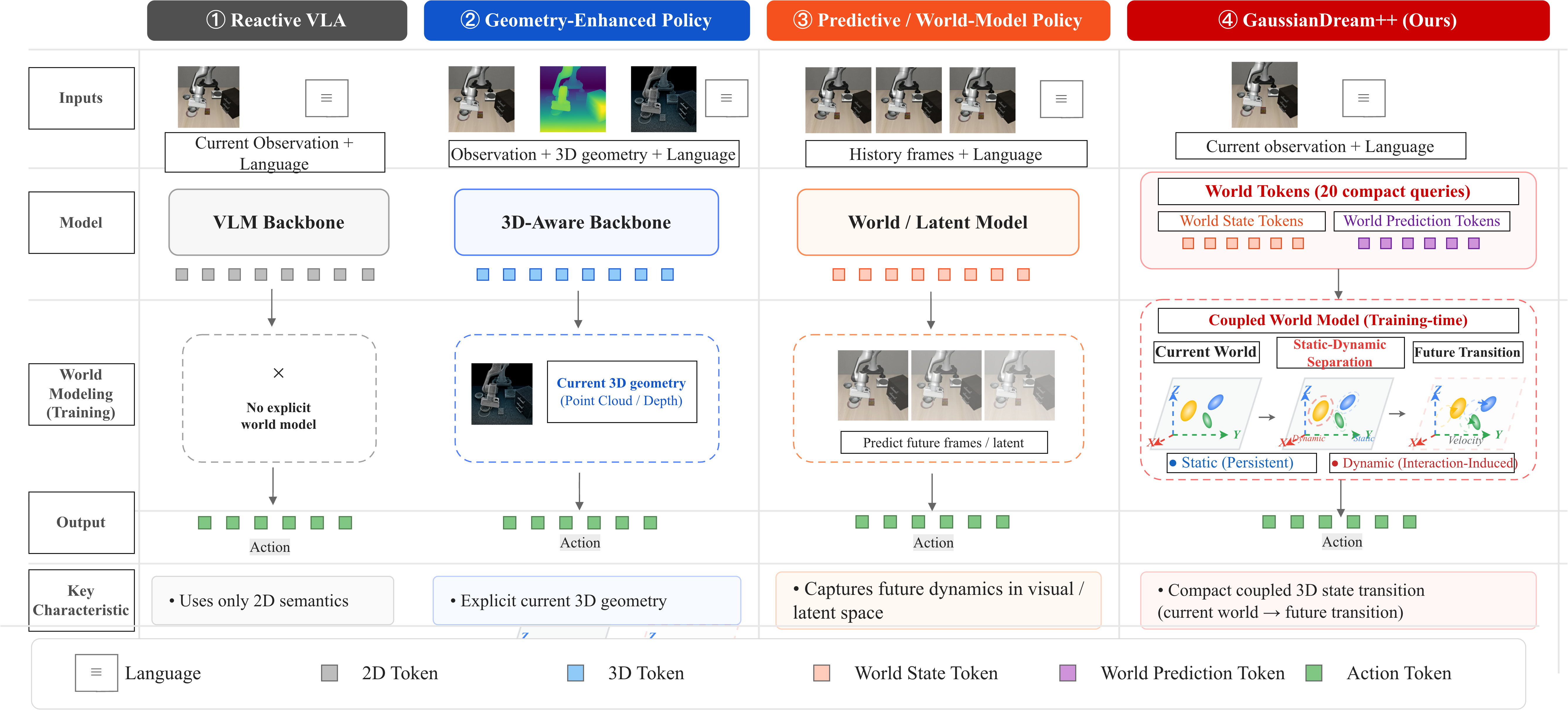}
    \caption{
    \textbf{Comparison of robotic policy paradigms.}
    Reactive VLAs map observations and language instructions directly to
    actions but rely mainly on implicit visual representations.
    Geometry-enhanced policies improve current-scene spatial grounding,
    whereas predictive and world-model policies introduce temporal
    supervision through future modeling.
    GaussianDream++ combines current-world reconstruction and future
    prediction within a compact, policy-native Gaussian representation,
    providing structured 3D supervision during training without requiring
    runtime world decoding or future rollout.
    }
    \label{fig:comparison}
\end{figure*}

Predictive geometry provides a promising bridge between explicit spatial
grounding and temporal modeling. GeoPredict introduces future kinematics and
predictive 3D Gaussian geometry as training-time supervision
~\citep{qian2025geopredict}. GaussianDream further shows that a learned
Gaussian prefix can be decoded into current and future Gaussian states,
turning ordinary robot trajectories into dense RGB, depth, and pseudo-3D
scene-flow supervision
~\citep{zhang2026gaussiandream}. Its asymmetric design removes the Gaussian
decoding heads during deployment and retains only the learned prefix for
action conditioning. This establishes an important principle: explicit 3D
world modeling can improve a feed-forward VLA policy without operating as an
online simulator.

GaussianDream nevertheless leaves room for a more compact and structured
world representation. Its prefix is constructed from temporal observations
through VGGT features, a Temporal Gaussian Evolution module, and a dense
$32\times32$ query grid. Although the resulting representation effectively
supports reconstruction, prediction, and action generation, online prefix
construction still requires the dedicated temporal geometry pathway.
Furthermore, the same 1024-token prefix jointly carries current geometry,
future evolution, and action-conditioning information. GaussianDream already
predicts future center displacements relative to a reconstructed Current
Gaussian template; the remaining question is therefore not whether residual
Gaussian prediction is useful, but whether current state and future change
can be organized into a substantially smaller, role-structured
representation that is native to the VLA policy. Explicitly separating
persistent structure from interaction-induced motion may further reduce
redundant temporal modeling and suppress spurious changes in static regions.

To this end, we introduce \textbf{GaussianDream++}, a compact, policy-native
upgrade of GaussianDream. GaussianDream++ replaces the dense external
Gaussian prefix with two complementary token groups embedded directly in the
PaliGemma backbone. \textbf{World State Tokens} represent the current
physical scene, while \textbf{World Prediction Tokens} represent
short-horizon evolution. These tokens are jointly processed with the original
visual-language context and remain directly visible to the Action Expert
through the native attention pathway. Consequently, the representation
shaped by Gaussian world supervision is also the representation used for
action generation, without an additional projection or action-conditioning
module.

During training, a lightweight \textbf{World Representation Head} decodes the
World State Tokens into a renderable Current World and combines them with the
World Prediction Tokens to model horizon-dependent Future Prediction over the
same Gaussian primitives. GaussianDream++ retains the primitive-aligned
residual formulation of GaussianDream, but reorganizes it through explicit
state/prediction roles and static--dynamic factorization. Persistent
primitives inherit the Current World structure, while residual motion is
concentrated on the robot, manipulated objects, and interaction regions.
This coupled formulation preserves current-to-future correspondence and uses
prediction capacity primarily for physically meaningful changes.

GaussianDream++ follows an asymmetric training-and-deployment strategy.
Gaussian reconstruction, differentiable rendering, metric depth, visibility,
and motion objectives provide dense physical supervision during learning.
At deployment, the World Representation Head, Gaussian renderer, and all
auxiliary supervision branches are removed, while the World State Tokens and
World Prediction Tokens remain in the VLA backbone. The deployed policy
therefore uses only 20 additional world tokens and requires neither the
runtime VGGT/TGE pathway used by GaussianDream, nor online Gaussian decoding,
rendering, or future rollout.

Experiments demonstrate that this compact formulation improves both
manipulation performance and distribution-shift robustness. Under the matched
GaussianDream-family protocol, GaussianDream++ achieves
\textbf{98.6\%} success on LIBERO and \textbf{87.8\%} on LIBERO-Plus,
outperforming GaussianDream by 0.8 percentage points overall. The gains are
particularly evident under Camera and Layout shifts, where GaussianDream++
improves GaussianDream by 2.8 and 1.6 points, respectively. Real-robot
experiments further improve average success from 29.2\% to 52.5\% over the
reproduced $\pi_{0.5}$ baseline across precision-demanding and long-horizon
manipulation tasks.

Our contributions are summarized as follows:

\begin{itemize}
    \item \textbf{Compact policy-native Gaussian world modeling.}
    We introduce GaussianDream++, which replaces GaussianDream's dense
    VGGT/TGE-based prefix with 20 World State and World Prediction Tokens
    embedded directly in the VLA backbone, allowing the Action Expert to use
    world-supervised representations through its native attention pathway.

    \item \textbf{Role-structured current and future representation.}
    World State Tokens reconstruct the Current World, while World Prediction
    Tokens represent short-horizon evolution over shared Gaussian primitives.
    Static--dynamic factorization preserves persistent scene structure and
    focuses residual prediction on interaction-induced changes.

    \item \textbf{Efficient asymmetric supervision and robust control.}
    Gaussian decoding and rendering are confined to training and removed at
    deployment, together with the dedicated runtime geometry pathway.
    Extensive simulation and real-robot evaluations demonstrate improved
    accuracy, robustness, and deployment efficiency over GaussianDream and
    the reproduced $\pi_{0.5}$ policy.
\end{itemize}
\section{Related Work}
\label{sec:related}

\subsection{Vision-Language-Action Policies and Geometric Grounding}

Vision-Language-Action (VLA) policies adapt pretrained vision-language
representations to robotic action generation, enabling language-conditioned
manipulation across diverse tasks and environments. Representative systems
including RT-2, OpenVLA, Octo, $\pi_0$, and $\pi_{0.5}$ demonstrate the
effectiveness of combining large-scale vision-language pretraining with robot
demonstrations
~\citep{brohan2023rt2,kim2024openvla,octo2024,black2024pi0,
physicalintelligence2025pi05}. Recent diffusion- and flow-matching-based
formulations further improve continuous action generation and closed-loop
control. Nevertheless, these policies are predominantly optimized through
action imitation, leaving metric scene geometry only indirectly constrained
by image features and action labels.

A growing body of work incorporates depth, stereo observations, point clouds,
spatial encodings, 4D features, or pretrained 3D representations into VLA
policies
~\citep{zhen20243d,sun2025geovla,li2025qdepthvla,
deng2025stereovla,zhou2025vla4d,abouzeid2025geoawarevla}.
Spatial Forcing aligns intermediate policy features with representations from
a pretrained spatial model
~\citep{li2025spatialforcing}, while G3VLA introduces calibrated geometric
inductive biases without changing the original action formulation
~\citep{peng2026g3vla}. PointACT studies how geometric features should
interact with action representations
~\citep{chen2026pointact}, and VGA explores vision-geometry backbones for
native spatial grounding
~\citep{song2026vga}. These methods establish the importance of explicit 3D
structure for precise manipulation, but primarily constrain the representation
of the current scene.

\subsection{Predictive Policies and World Models}

Predictive policies and World-Action Models introduce future observations,
trajectories, video representations, or latent dynamics as additional
supervision beyond one-step behavior cloning
~\citep{cen2025worldvla,bi2025motus,kim2026cosmospolicy,
ye2026dreamzero,li2026lingbotva}. By exposing the policy to scene evolution,
these methods provide useful temporal priors for action learning and planning.
However, future states are commonly modeled in RGB or video-latent spaces,
where visual coherence does not necessarily imply metrically consistent
physical change, particularly under contact, occlusion, and precise object
placement.

Maintaining a large generative model or explicit future rollout during
deployment can also increase inference cost. Fast-WAM shows that predictive
co-training can improve action learning without test-time future imagination
~\citep{yuan2026fast}, motivating asymmetric designs that use rich predictive
objectives during training while keeping deployment lightweight. SwiftVLA
incorporates spatio-temporal geometry and future-trajectory supervision
~\citep{ni2025swiftvla}; LaMP uses 3D scene flow as a latent motion prior
~\citep{wang2026lamp}; and GeoPredict combines future kinematics with
predictive 3D Gaussian geometry
~\citep{qian2025geopredict}. These approaches demonstrate the value of
structured future supervision, while leaving open how current geometry and
future evolution should be represented compactly inside an efficient VLA
policy.

\subsection{Gaussian World Modeling for Robot Learning}

3D Gaussian Splatting represents a scene with explicit primitives carrying
position, scale, rotation, opacity, and appearance, while supporting efficient
differentiable rendering
~\citep{kerbl20233d}. This primitive-based formulation provides a natural
interface for associating geometry, visibility, appearance, and motion with
corresponding physical scene elements. Gaussian representations have
subsequently been applied to robotic reconstruction and manipulation.
ManiGaussian introduces dynamic Gaussian representations for multi-task
robot control, while GWM studies scalable Gaussian world models for robotic
manipulation
~\citep{lu2024manigaussian,lu2025gwm}.

The closest predecessor to our work is GaussianDream
~\citep{zhang2026gaussiandream}. It constructs a GaussianDream prefix from
temporal geometric features and uses training-only reconstruction and
prediction heads to produce current and future Gaussian states. RGB, depth,
and pseudo-3D scene-flow objectives convert robot trajectories into dense
spatio-temporal supervision, while the auxiliary Gaussian heads are removed
at deployment. Its future branch already predicts horizon-conditioned center
displacements over the reconstructed current Gaussian template.

GaussianDream++ retains this training-time supervision principle but revisits
the construction and organization of the policy representation. Rather than
relying on a dedicated dense temporal-geometry prefix, it embeds a compact,
role-structured world representation directly in the VLA backbone. The
technical realization of this policy-native representation and its
current/future supervision is described in Sec.~\ref{sec:method}.
\section{Method}
\label{sec:method}

\begin{figure*}[t]
    \centering
    \includegraphics[width=0.98\textwidth]
    {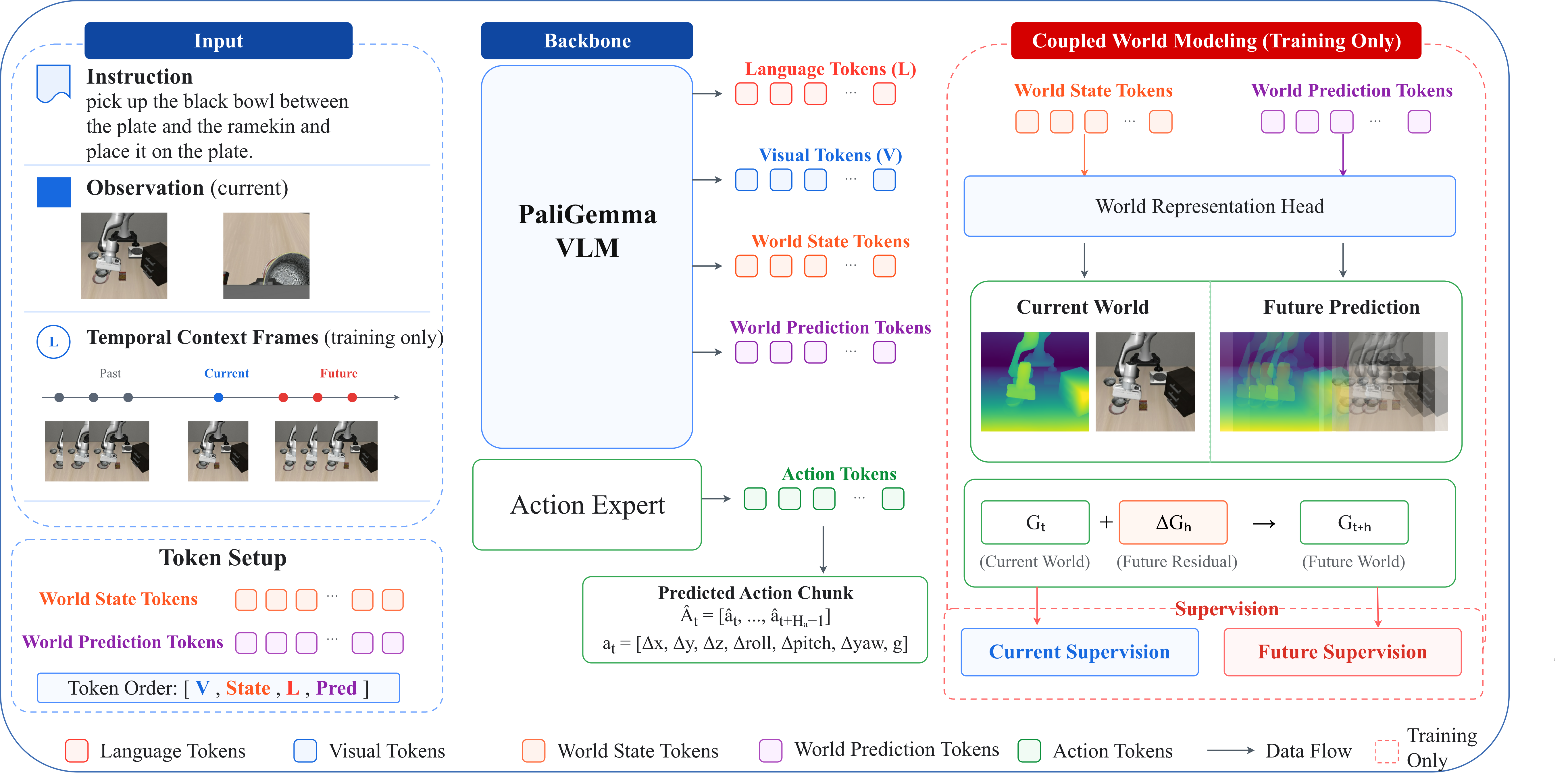}
    \caption{
    \textbf{Overview of GaussianDream++.}
    World State Tokens and World Prediction Tokens are inserted directly into
    the PaliGemma prefix and contextualized with visual and language tokens.
    The resulting representation directly conditions the Action Expert.
    During training, a lightweight World Representation Head decodes the same
    token representations into the Current World and coupled Future Prediction
    for dense Gaussian supervision. Future observations are used only to
    construct supervision targets and never enter the policy forward pass.
    At deployment, the World Representation Head, Gaussian renderer, and all
    auxiliary supervision branches are removed, while the world tokens remain
    in the policy.
    }
    \label{fig:framework}
\end{figure*}

\subsection{Overview}
\label{sec:overview}

GaussianDream++ retains the training-time Gaussian supervision principle of
GaussianDream~\citep{zhang2026gaussiandream}, but moves the supervised world
representation from a dedicated temporal geometry pathway into the native
VLA backbone. As shown in Fig.~\ref{fig:framework}, the current multi-view
observation and language instruction are processed by PaliGemma together with
two groups of learnable tokens: \textbf{World State Tokens}, which represent
the current physical scene, and \textbf{World Prediction Tokens}, which
represent its short-horizon evolution. Their contextualized hidden states
form part of the same multimodal prefix accessed by the flow-matching Action
Expert.

The same token representations are decoded during training by a lightweight
\textbf{World Representation Head}. The state representation is constrained
through Current World reconstruction, whereas the prediction representation
is constrained through Future Prediction over the same Gaussian primitives.
This auxiliary decoding path converts robot trajectories into dense
supervision for geometry, appearance, visibility, and motion. It does not
provide additional policy inputs: future observations are used exclusively
as supervision targets, and neither rendered images nor explicit Gaussian
primitives are consumed by the Action Expert.

The training and deployment paths are asymmetric. After training, the World
Representation Head, Gaussian renderer, target-construction modules, and
auxiliary objectives are discarded. World State Tokens and World Prediction
Tokens remain in the PaliGemma prefix and continue to condition action
generation through the native VLA pathway. GaussianDream++ therefore retains
structured current-and-future world supervision without online Gaussian
decoding, rendering, future rollout, or the dedicated runtime VGGT/TGE
prefix-construction path used by GaussianDream.

\subsection{Policy-Native World Representation}
\label{sec:world_representation}

Let $o_t$ denote the current multi-view observation, $\ell$ the language
instruction, and $r_t$ the robot state. The policy predicts an action chunk
$A_t=[a_t,\ldots,a_{t+H_a-1}]$. GaussianDream++ augments the multimodal
prefix with two learnable token sets,
\begin{equation}
    Z^{S}\in\mathbb{R}^{N_s\times d},
    \qquad
    Z^{P}\in\mathbb{R}^{N_p\times d},
    \label{eq:world_tokens}
\end{equation}
where $Z^{S}$ and $Z^{P}$ denote World State Tokens and World Prediction
Tokens, respectively. In the main configuration, we use $N_s=16$ state
tokens and $N_p=4$ prediction tokens. The state tokens retain a
$4\times4$ coarse spatial organization and receive calibrated spatial and
ray embeddings, while the prediction tokens are associated with
short-horizon prediction slots. These tokens are compact latent summaries
rather than one-to-one Gaussian primitives.

PaliGemma jointly contextualizes the world tokens with the original visual
and language tokens:
\begin{equation}
    \left(
        H_t^{VL},
        H_t^{S},
        H_t^{P}
    \right)
    =
    F_{\theta}
    \left(
        o_t,\ell;
        Z^{S},Z^{P}
    \right),
    \label{eq:world_backbone}
\end{equation}
where $H_t^{VL}$ denotes the contextualized visual-language representation,
and $H_t^{S}$ and $H_t^{P}$ are the final hidden states of the World State
Tokens and World Prediction Tokens.

The Action Expert directly conditions on the complete world-token-augmented
prefix:
\begin{equation}
    A_t
    =
    \pi_{\phi}
    \left(
        r_t;
        H_t^{VL},
        H_t^{S},
        H_t^{P}
    \right).
    \label{eq:world_conditioned_action}
\end{equation}
No additional world-to-action projector is introduced. Consequently, the
representations constrained by Gaussian reconstruction and prediction are
also directly available to action generation.

During training, the World Representation Head reads the same hidden states
to produce explicit Gaussian supervision:
\begin{equation}
    \left(
        G_t,
        \left\{
            \widehat{G}_{t+h}
        \right\}_{h\in\mathcal{H}}
    \right)
    =
    W_{\psi}
    \left(
        H_t^{S},
        H_t^{P}
    \right),
    \label{eq:world_head}
\end{equation}
where $G_t$ is the reconstructed Current World and
$\widehat{G}_{t+h}$ denotes the predicted Gaussian state at horizon $h$.
The explicit worlds supervise the policy representation but are never fed
back into the Action Expert. This separates the compact representation
carried by the deployed policy from the dense Gaussian outputs used only for
training.

Compared with the unified 1024-query GaussianDream prefix, GaussianDream++
uses only 20 role-specific tokens formed directly inside PaliGemma. The
result is a substantially smaller world representation with explicit
current-state and future-prediction roles, while the details of Current World
reconstruction and coupled Future Prediction are introduced in the following
subsections.

\subsection{Current World Reconstruction}
\label{sec:current_world}

The World State Tokens are constrained to encode a spatially grounded
representation of the current environment. Their contextualized hidden states
$H_t^{S}$ retain a coarse spatial organization and are expanded by the
World Representation Head into a dense Gaussian feature field,
\begin{equation}
    F_t^{G}
    =
    D_{\mathrm{state}}
    \left(
        \operatorname{Grid}(H_t^{S})
    \right),
    \label{eq:state_decode}
\end{equation}
where $D_{\mathrm{state}}$ denotes the state-decoding branch of the World
Representation Head. The compact World State Tokens therefore serve as
latent summaries rather than being placed in one-to-one correspondence with
Gaussian primitives.

\paragraph{Geometry--appearance decomposition.}
We decode geometry and appearance through separate prediction branches. The
geometry branch predicts metric depth and non-appearance Gaussian attributes,
while the appearance branch operates on stop-gradient geometry features with
optional current-image feature fusion:
\begin{align}
    \bar D_t,\Theta_t^{\mathrm{geo}},F_t^{\mathrm{geo}}
    &=
    H_{\mathrm{geo}}(F_t^{G}), \nonumber\\
    \Theta_t^{\mathrm{app}}
    &=
    H_{\mathrm{app}}
    \left(
        \operatorname{sg}[F_t^{\mathrm{geo}}],
        f_t^{\mathrm{img}}
    \right).
    \label{eq:gaussian_attributes}
\end{align}
Here, $\Theta_t^{\mathrm{geo}}$ contains scale, rotation, and opacity, whereas
$\Theta_t^{\mathrm{app}}$ contains the spherical-harmonic appearance
coefficients. Detaching the geometry features before appearance prediction
prevents photometric shortcuts from altering the metric scene structure.

Gaussian centers are obtained by unprojecting the predicted metric depth
using the calibrated camera parameters,
\begin{equation}
    \{\mu_i^t\}_{i=1}^{N_G}
    =
    \mathcal{U}
    \left(
        \bar D_t;K_t,T_t
    \right),
    \label{eq:depth_unprojection}
\end{equation}
where $\mathcal{U}$ denotes multi-view metric unprojection, and $K_t$ and
$T_t$ are the camera intrinsics and extrinsics. The resulting Current World
is
\begin{equation}
    G_t
    =
    \left\{
    \left(
        \mu_i^t,
        s_i^t,
        q_i^t,
        \alpha_i^t,
        c_i^t
    \right)
    \right\}_{i=1}^{N_G}.
    \label{eq:current_world}
\end{equation}

The Current World is rendered through differentiable Gaussian splatting,
\begin{equation}
    \left(
        \hat I_t,
        \hat D_t,
        \hat A_t
    \right)
    =
    \mathcal{R}
    \left(
        G_t;K_t,T_t
    \right),
    \label{eq:current_render}
\end{equation}
where $\hat I_t$, $\hat D_t$, and $\hat A_t$ denote rendered RGB, metric
depth, and accumulated Gaussian coverage, respectively. We supervise the
Current World with
\begin{equation}
    \mathcal{L}_{\mathrm{cur}}
    =
    \lambda_{\mathrm{rgb}}\mathcal{L}_{\mathrm{rgb}}
    +
    \lambda_{\mathrm{depth}}\mathcal{L}_{\mathrm{depth}}
    +
    \lambda_{\alpha}\mathcal{L}_{\alpha}
    +
    \lambda_{\mathrm{reg}}\mathcal{L}_{\mathrm{reg}}.
    \label{eq:current_objective}
\end{equation}
The RGB term combines photometric and structural similarity objectives,
whereas depth and alpha losses are evaluated only over valid supervision
regions. Missing depth is excluded rather than interpreted as transparent
background. Current World reconstruction is not intended as a
photorealistic reconstruction task; it provides explicit constraints on
metric geometry, visibility, and object layout for the representations used
by the Action Expert.

\subsection{Coupled Future Prediction}
\label{sec:future_prediction}

Current World reconstruction anchors the policy representation to the scene
at time $t$, but manipulation additionally requires short-horizon
anticipation. Following the primitive-aligned formulation established by
GaussianDream, GaussianDream++ preserves the reconstructed Current World as a
shared template and predicts only horizon-dependent geometric changes.
Its distinction lies in decoding these changes from the role-specific World
Prediction Tokens rather than from a single dense prefix.

For a prediction horizon $h\in\mathcal{H}$, the corresponding World
Prediction representation $H_{t,h}^{P}$ is combined with the Current World
features and a learnable horizon embedding $e_h$:
\begin{equation}
    \Delta\mu_i^{h}
    =
    H_{\Delta}
    \left(
        F_{t,i}^{G},
        H_{t,h}^{P},
        e_h
    \right).
    \label{eq:future_residual}
\end{equation}
When static--dynamic gating is enabled, the transition head additionally
predicts a motion coefficient $m_i^{h}\in[0,1]$; otherwise
$m_i^{h}=1$. The future center is composed as
\begin{equation}
    \mu_i^{t+h}
    =
    \mu_i^t
    +
    m_i^{h}\Delta\mu_i^{h}.
    \label{eq:future_center}
\end{equation}
Scale, rotation, opacity, and appearance are inherited from the Current
World, yielding
\begin{equation}
    G_{t+h}
    =
    \left\{
    \left(
        \mu_i^{t+h},
        s_i^t,
        q_i^t,
        \alpha_i^t,
        c_i^t
    \right)
    \right\}_{i=1}^{N_G}.
    \label{eq:future_world}
\end{equation}
This shared-primitive construction preserves current-to-future
correspondence and allocates predictive capacity to geometric change rather
than repeatedly regenerating persistent scene attributes. Static-consistency
supervision further penalizes residual motion in regions identified as
persistent by valid motion targets, reducing spurious drift in the
background and stationary objects.

Each Future World is rendered using the camera parameters associated with
its prediction horizon,
\begin{equation}
    \left(
        \hat I_{t+h},
        \hat D_{t+h},
        \hat A_{t+h}
    \right)
    =
    \mathcal{R}
    \left(
        G_{t+h};K_{t+h},T_{t+h}
    \right).
    \label{eq:future_render}
\end{equation}
Using horizon-specific calibration prevents changes in viewpoint from being
absorbed into the predicted physical motion. Future supervision combines
rendering, metric depth, three-dimensional motion, and static-consistency
terms:
\begin{equation}
    \mathcal{L}_{\mathrm{fut}}
    =
    \sum_{h\in\mathcal{H}}
    w_h
    \left(
        \lambda_{\mathrm{rgb}}
        \mathcal{L}_{\mathrm{rgb}}^{h}
        +
        \lambda_{\mathrm{depth}}
        \mathcal{L}_{\mathrm{depth}}^{h}
        +
        \lambda_{\mathrm{flow}}
        \mathcal{L}_{\mathrm{flow}}^{h}
        +
        \lambda_{\mathrm{stat}}
        \mathcal{L}_{\mathrm{stat}}^{h}
    \right).
    \label{eq:future_objective}
\end{equation}
All terms are computed only for valid horizons and valid geometric
correspondences. Future observations supply the targets in
Eq.~\ref{eq:future_objective} but never enter the policy forward pass.
Moreover, because World Prediction Tokens are conditioned on observations
and task context rather than candidate actions, the model learns anticipated
short-horizon evolution rather than counterfactual action-conditioned
simulation.

\subsection{World-Action Learning and Efficient Deployment}
\label{sec:learning_deployment}

GaussianDream++ jointly optimizes action generation and Gaussian world
supervision so that the policy-native World Tokens remain both physically
structured and useful for control. Let $\mathcal{L}_{\mathrm{act}}$ denote
the original flow-matching action objective of the $\pi_{0.5}$ policy. The
complete objective is
\begin{equation}
    \mathcal{L}
    =
    \mathcal{L}_{\mathrm{act}}
    +
    \lambda_{\mathrm{cur}}
    \mathcal{L}_{\mathrm{cur}}
    +
    \lambda_{\mathrm{fut}}
    \mathcal{L}_{\mathrm{fut}}.
    \label{eq:joint_objective}
\end{equation}
The action loss aligns the world-token-augmented prefix with executable
action chunks, while the Current World and Future Prediction losses prevent
the same representation from collapsing into unconstrained action-only
features. The world objectives remain active during policy fine-tuning,
ensuring that physical structure is retained as the Action Expert adapts.

The training and deployment paths are deliberately asymmetric. During
training, the World Representation Head, Gaussian renderer, future-target
builders, and all auxiliary objectives transform robot trajectories into
dense geometric and temporal supervision. At deployment, these components
are removed. The policy retains the PaliGemma backbone, World State Tokens,
World Prediction Tokens, the action adapters used during policy learning,
and the original Action Expert. Consequently, inference requires neither
the VGGT/TGE prefix-construction pathway of GaussianDream nor online Gaussian
reconstruction, rendering, or future rollout; the only additional policy
computation is the processing of the compact World Tokens through the native
VLA backbone.
\section{Experiments}
\label{sec:experiments}

We evaluate \textbf{GaussianDream++} from four perspectives:
\textbf{(1)} manipulation performance on standard tasks;
\textbf{(2)} zero-shot robustness under spatial, visual, linguistic, and
embodiment shifts;
\textbf{(3)} transfer to real-world manipulation; and
\textbf{(4)} the contributions and computational costs of its principal
design choices.

\subsection{Experimental Setup}
\label{sec:setup}

\paragraph{Benchmarks and metrics.}
We evaluate GaussianDream++ on LIBERO~\citep{liu2023libero} and
LIBERO-Plus~\citep{fei2026liberoplus}.
LIBERO contains four task suites---Spatial, Object, Goal, and Long---covering
spatial reasoning, object-centric manipulation, goal-conditioned control, and
long-horizon execution.
We report the average task success rate across the four suites.

LIBERO-Plus evaluates zero-shot robustness under seven controlled
distribution shifts:
\emph{Camera}, \emph{Robot}, \emph{Language}, \emph{Lighting},
\emph{Background}, \emph{Noise}, and \emph{Layout}.
Following the official protocol, policies are trained on standard LIBERO and
evaluated directly on LIBERO-Plus without robustness-specific adaptation.
The LIBERO-Plus Overall score follows the official aggregation over all
evaluation instances and is therefore not necessarily equal to the
unweighted mean of the seven shift categories.

\paragraph{Baselines and evaluation protocols.}
We compare against representative VLA, geometry-enhanced, predictive, and
world-model policies.
For standard LIBERO, we include GeoPredict~\citep{qian2025geopredict},
QDepth-VLA~\citep{li2025qdepthvla},
LingBot-VA~\citep{li2026lingbotva},
GeoVLA~\citep{sun2025geovla},
VLA-4D~\citep{zhou2025vla4d}, and
Spatial Forcing~\citep{li2025spatialforcing}.
For LIBERO-Plus, we report published zero-shot results from
OpenVLA-OFT~\citep{kim2025openvlaoft},
RIPT-VLA~\citep{tan2025riptvla},
StarVLA-$\alpha$~\citep{ye2026starvlaalpha},
Cosmos Policy~\citep{kim2026cosmospolicy},
LaMP~\citep{wang2026lamp},
$\pi_{0.5}$~\citep{physicalintelligence2025pi05}, and
ACoT-VLA~\citep{zhong2026acotvla}.
For ACoT-VLA, we use its zero-shot result rather than its separately reported
supervised fine-tuning result to maintain protocol consistency.

Published results may differ in training data, checkpoint selection, and
implementation details; they are included to contextualize overall
performance rather than support strict pairwise claims.
Our direct comparisons use the final block of
Tab.~\ref{tab:main_results}, where reproduced $\pi_{0.5}$,
GaussianDream, and GaussianDream++ are evaluated under the same
GaussianDream-family protocol.

\paragraph{Implementation.}
GaussianDream++ adopts the PaliGemma/$\pi_{0.5}$ backbone and its
flow-matching Action Expert.
The policy introduces 16 World State Tokens and 4 World Prediction Tokens
directly into the PaliGemma prefix.
During training, the action objective is jointly optimized with Current World
and Future Prediction supervision produced by the World Representation Head.
Future observations are used only to construct supervision targets and never
enter the policy forward pass.

At deployment, the World Representation Head, Gaussian renderer, and all
auxiliary supervision branches are removed.
World State Tokens and World Prediction Tokens remain in the VLA prefix and
condition the Action Expert through the native attention pathway.
Unlike GaussianDream, GaussianDream++ does not require the dedicated runtime
VGGT/TGE pathway for world-prefix construction.

\subsection{Main Results on LIBERO and LIBERO-Plus}
\label{sec:main_results}

\begin{table*}[t]
    \centering
    \caption{
    \textbf{Main results on LIBERO and LIBERO-Plus.}
    We report task success rate (\%).
    \textbf{(a)} Standard LIBERO results across the four task suites.
    \textbf{(b)} Zero-shot robustness on LIBERO-Plus.
    Published results are included for contextual comparison, whereas direct
    improvement claims are based on the matched GaussianDream-family blocks.
    LIBERO-Plus Overall follows the official task-level aggregation rather
    than the unweighted mean of the seven shifts.
    Best results within each evaluation block are shown in \textbf{bold}.
    }
    \label{tab:main_results}

    \small
    \setlength{\tabcolsep}{8pt}

    \begin{tabular*}{\textwidth}{
        @{\extracolsep{\fill}}
        lccccc
    }
        \toprule
        \multicolumn{6}{c}{
            \textbf{(a) Standard LIBERO}
        } \\
        \midrule
        Method
        & Spatial
        & Object
        & Goal
        & Long
        & Average \\
        \midrule

        \multicolumn{6}{c}{
            \textit{Published results}
        } \\
        \midrule

        $\pi_0$~\citep{black2024pi0}
        & 96.8 & 98.8 & 95.8 & 85.2 & 94.1 \\

        $\pi_{0.5}$~\citep{physicalintelligence2025pi05}
        & 97.8 & 98.8 & 97.6 & 92.4 & 96.7 \\

        GeoPredict~\citep{qian2025geopredict}
        & 98.0 & 98.2 & 95.7 & 94.0 & 96.5 \\

        QDepth-VLA~\citep{li2025qdepthvla}
        & 97.6 & 96.6 & 95.2 & 90.0 & 94.9 \\

        LingBot-VA~\citep{li2026lingbotva}
        & 98.5 & 99.6 & 97.2 & \textbf{98.5} & \textbf{98.5} \\

        GeoVLA~\citep{sun2025geovla}
        & 98.4 & 99.0 & 96.6 & 96.6 & 97.7 \\

        VLA-4D~\citep{zhou2025vla4d}
        & 97.9 & 98.6 & 97.8 & 94.8 & 97.4 \\

        3D-CAVLA~\citep{bhat20253dcavla}
        & 98.2 & \textbf{99.8} & \textbf{98.2} & 96.1 & 98.1 \\

        Spatial Forcing (PyTorch)~\citep{li2025spatialforcing}
        & \textbf{98.6} & 98.4 & \textbf{98.2} & 95.4 & 97.6 \\

        \midrule
        \multicolumn{6}{c}{
            \textit{Matched GaussianDream-family protocol}
        } \\
        \midrule

        $\pi_{0.5}$ reproduced
        & 98.8 & 98.2 & 98.0 & 92.4 & 96.9 \\

        GaussianDream~\citep{zhang2026gaussiandream}
        & 99.0 & \textbf{99.6} & \textbf{99.0} & 96.0 & 98.4 \\

        \rowcolor{gray!12}
        \textbf{GaussianDream++ (Ours)}
        & \textbf{99.2}
        & \textbf{99.6}
        & \textbf{99.0}
        & \textbf{96.6}
        & \textbf{98.6} \\

        \bottomrule
    \end{tabular*}

    \vspace{5pt}

    \scriptsize
    \setlength{\tabcolsep}{3.6pt}

    \begin{tabular*}{\textwidth}{
        @{\extracolsep{\fill}}
        lcccccccc
    }
        \toprule
        \multicolumn{9}{c}{
            \textbf{(b) LIBERO-Plus Zero-Shot Robustness}
        } \\
        \midrule

        Method
        & Camera
        & Robot
        & Lang.
        & Light
        & BG
        & Noise
        & Layout
        & Overall \\
        \midrule

        \multicolumn{9}{c}{
            \textit{Published zero-shot results}
        } \\
        \midrule

        OpenVLA-OFT~\citep{kim2025openvlaoft}
        & 56.4 & 31.9 & 79.5 & 88.7
        & 93.3 & 75.8 & 74.2 & 69.6 \\

        RIPT-VLA~\citep{tan2025riptvla}
        & 55.2 & 31.2 & 77.6 & 88.4
        & 91.6 & 73.5 & 74.2 & 68.4 \\

        StarVLA-$\alpha$ (Specialist)~\citep{ye2026starvlaalpha}
        & 48.7 & 63.4 & 86.8 & 95.8
        & 94.6 & 75.0 & 80.2 & 77.8 \\

        StarVLA-$\alpha$ (Generalist)~\citep{ye2026starvlaalpha}
        & 52.5 & 64.3 & 86.2 & \textbf{97.8}
        & \textbf{98.1} & 80.2 & 79.1 & 79.7 \\

        Cosmos Policy~\citep{kim2026cosmospolicy}
        & \textbf{75.8} & 63.3 & 81.7 & 96.5
        & 88.9 & \textbf{92.7} & 82.2 & 82.2 \\

        LaMP~\citep{wang2026lamp}
        & 64.5 & 69.6 & \textbf{88.2} & 95.3
        & 97.4 & 76.9 & 73.8 & 79.3 \\

        $\pi_{0.5}$ (ACoT protocol)~
        \citep{physicalintelligence2025pi05}
        & \textbf{75.8} & 79.4 & 83.3 & 95.5
        & 95.0 & 89.6 & 87.0 & 85.7 \\

        ACoT-VLA~\citep{zhong2026acotvla}
        & 72.6 & \textbf{82.6} & 87.5 & 97.7
        & 96.5 & 87.8 & \textbf{88.1} & \textbf{86.6} \\

        \midrule
        \multicolumn{9}{c}{
            \textit{Matched GaussianDream-family protocol}
        } \\
        \midrule

        $\pi_{0.5}$ reproduced
        & 73.2 & \textbf{77.9} & 84.3 & 96.3
        & 95.5 & 89.9 & 87.7 & 85.5 \\

        GaussianDream~\citep{zhang2026gaussiandream}
        & 77.3 & 72.1 & 86.9 & 98.9
        & 98.5 & 93.6 & 88.4 & 87.0 \\

        \rowcolor{gray!12}
        \textbf{GaussianDream++ (Ours)}
        & \textbf{80.1}
        & 73.0
        & \textbf{87.4}
        & \textbf{99.1}
        & \textbf{98.8}
        & \textbf{94.2}
        & \textbf{90.0}
        & \textbf{87.8} \\

        \bottomrule
    \end{tabular*}
\end{table*}

\paragraph{Overall performance.}
GaussianDream++ achieves 98.6\% average success on LIBERO and 87.8\%
Overall on LIBERO-Plus.
Standard LIBERO is highly saturated, yet GaussianDream++ remains competitive
with the strongest published policies and improves the matched GaussianDream
and reproduced $\pi_{0.5}$ baselines by 0.2 and 1.7 percentage points,
respectively.
On LIBERO-Plus, GaussianDream++ obtains the highest Overall score in
Tab.~\ref{tab:main_results}.
Because independently reported results may use different training and
evaluation implementations, we base quantitative improvement claims on the
matched GaussianDream-family block: GaussianDream++ improves Overall by
2.3 points over reproduced $\pi_{0.5}$ and by 0.8 points over GaussianDream.

\paragraph{Robustness under geometry-sensitive shifts.}
The clearest gains occur under Camera and Layout perturbations.
Compared with reproduced $\pi_{0.5}$, GaussianDream++ improves Camera from
73.2\% to 80.1\% and Layout from 87.7\% to 90.0\%, corresponding to gains of
6.9 and 2.3 points.
Relative to GaussianDream, the corresponding gains are 2.8 and 1.6 points.
Camera shifts alter the projection of an otherwise similar physical scene,
whereas Layout shifts change object-relative spatial relationships.
The improvements are therefore consistent with the intended role of Current
World supervision and policy-native world tokens: the policy is encouraged to
retain scene geometry and object layout rather than depend solely on
viewpoint-specific appearance.

GaussianDream++ also improves Noise from 89.9\% to 94.2\% over reproduced
$\pi_{0.5}$, suggesting that dense Gaussian supervision provides a useful
structural constraint when local image evidence is degraded.
Performance under Robot shift, however, remains below reproduced
$\pi_{0.5}$.
This indicates that the present gains are strongest for viewpoint, layout,
and visual perturbations, while embodiment-level generalization remains a
separate challenge.

\paragraph{GaussianDream-family comparison.}
GaussianDream validates the effectiveness of current and future Gaussian
supervision, while GaussianDream++ reorganizes the same supervision into
role-separated World State Tokens and World Prediction Tokens embedded
directly in the VLA backbone.
The matched gains show that replacing the dense VGGT/TGE-based prefix with a
compact policy-native representation does not sacrifice manipulation
accuracy.
Instead, GaussianDream++ improves the distribution-shift dimensions most
closely associated with geometric consistency, while removing the dedicated
runtime geometry pathway.

\subsection{Real-World Manipulation}
\label{sec:real_world}

We evaluate GaussianDream++ on a dual-arm physical robot platform to examine
whether its policy-native world representation transfers beyond simulation.
The evaluation complements the Camera and Layout shifts in LIBERO-Plus with
two tasks that emphasize complementary aspects of physical manipulation.
\emph{Bowl-Proximity} requires moving one white bowl adjacent to another
while keeping both bowls upright, and therefore stresses object-relative
localization and precision placement.
\emph{Eggplant-to-Pink-Plate} requires grasping the eggplant and placing it
inside the pink plate in the presence of a bitter melon and a blue plate,
testing semantic disambiguation, distractor rejection, and long-horizon
execution.

Each task is evaluated under three conditions.
\textbf{Standard} uses the nominal object arrangement and global-camera pose;
\textbf{Layout} changes the relative configuration of task-relevant objects
and distractors while preserving the instruction; and
\textbf{Camera} physically relocates the global camera while keeping the
workspace and manipulation objective unchanged.
The wrist-mounted camera remains fixed, isolating the effect of the global
viewpoint change.
For Bowl-Proximity, a rollout is successful when the commanded adjacency is
established while both bowls remain upright.
For Eggplant-to-Pink-Plate, success requires the eggplant to be released
inside the target plate without human intervention.
We evaluate each method for 20 trials per task and condition using the same
task definitions and success criteria.

\begin{figure*}[t]
    \centering
    \includegraphics[width=0.98\textwidth]{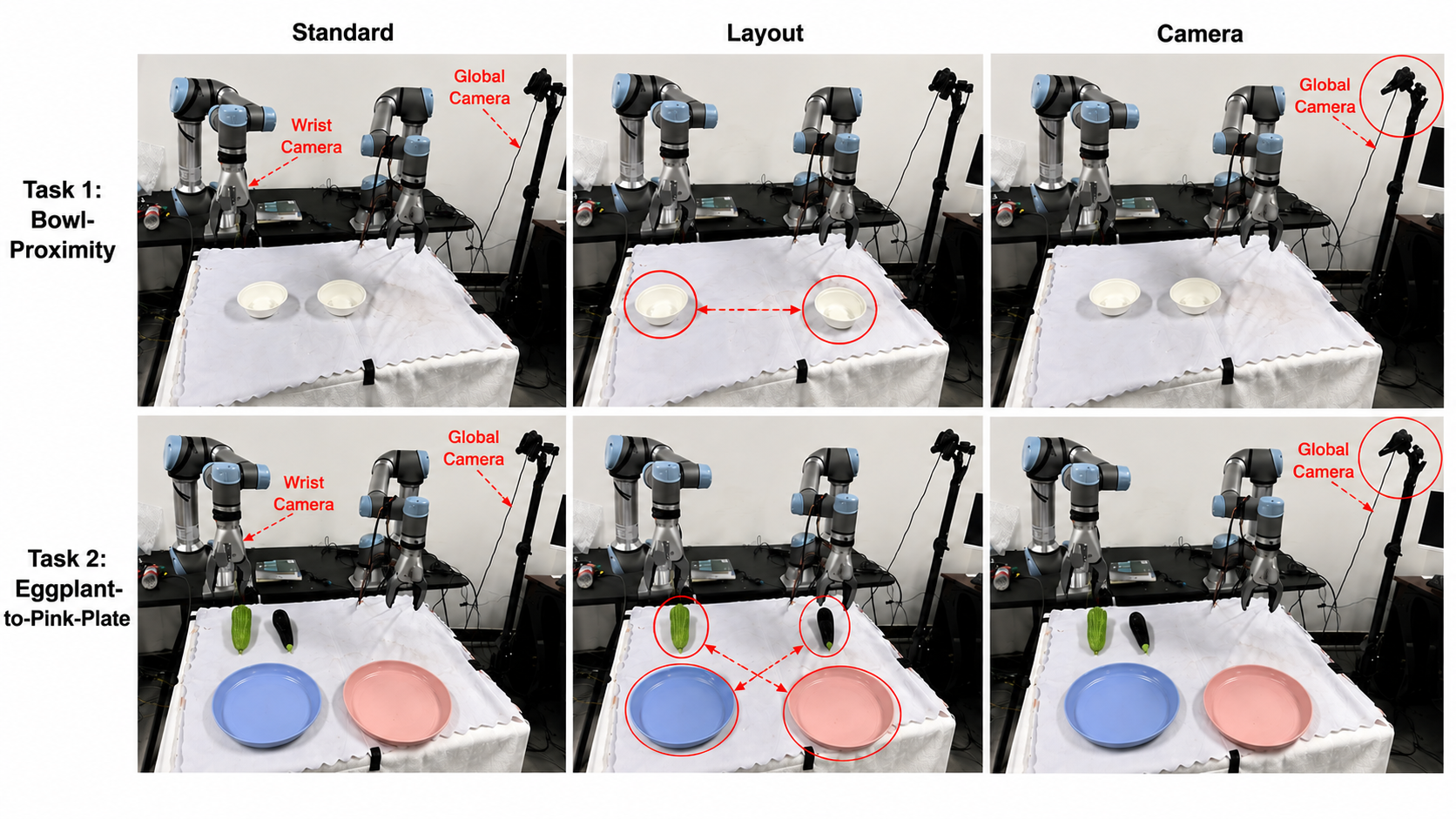}
    \caption{
    \textbf{Real-world tasks and distribution shifts.}
    Bowl-Proximity evaluates fine-grained object-relative placement, whereas
    Eggplant-to-Pink-Plate evaluates long-horizon manipulation with semantic
    distractors. Each task is tested under the nominal Standard setting, a
    Layout shift that changes object-relative configurations, and a Camera
    shift that relocates the global camera while retaining the same task goal.
    }
    \label{fig:real_world}
\end{figure*}

\begin{table*}[t]
    \centering
    \caption{
    \textbf{Real-world manipulation results.}
    Each task--condition pair contains 20 trials.
    Entries report successful trials followed by success rate (\%) in
    parentheses. The Overall column pools the three conditions for each task,
    while the final group pools both tasks.
    }
    \label{tab:real_world}

    \small
    \renewcommand{\arraystretch}{1.12}
    \setlength{\tabcolsep}{7pt}

    \begin{tabular*}{0.96\textwidth}{
        @{\extracolsep{\fill}}
        lcccc
        @{}
    }
        \toprule
        Method
        & Standard
        & Layout
        & Camera
        & Overall \\
        \midrule

        \multicolumn{5}{@{}l}{
            \textit{Task 1: Bowl-Proximity}
        } \\[-1pt]

        $\pi_{0.5}$ reproduced
        & 7/20 (35.0)
        & 4/20 (20.0)
        & 4/20 (20.0)
        & 15/60 (25.0) \\

        \rowcolor{gray!10}
        \textbf{GaussianDream++}
        & 11/20 (55.0)
        & 10/20 (50.0)
        & 7/20 (35.0)
        & \textbf{28/60 (46.7)} \\

        \midrule

        \multicolumn{5}{@{}l}{
            \textit{Task 2: Eggplant-to-Pink-Plate}
        } \\[-1pt]

        $\pi_{0.5}$ reproduced
        & 9/20 (45.0)
        & 6/20 (30.0)
        & 5/20 (25.0)
        & 20/60 (33.3) \\

        \rowcolor{gray!10}
        \textbf{GaussianDream++}
        & 14/20 (70.0)
        & 11/20 (55.0)
        & 10/20 (50.0)
        & \textbf{35/60 (58.3)} \\

        \midrule

        \multicolumn{5}{@{}l}{
            \textit{All tasks}
        } \\[-1pt]

        $\pi_{0.5}$ reproduced
        & 16/40 (40.0)
        & 10/40 (25.0)
        & 9/40 (22.5)
        & 35/120 (29.2) \\

        \rowcolor{gray!10}
        \textbf{GaussianDream++}
        & 25/40 (62.5)
        & 21/40 (52.5)
        & 17/40 (42.5)
        & \textbf{63/120 (52.5)} \\

        \bottomrule
    \end{tabular*}
\end{table*}

GaussianDream++ improves the pooled success rate from 29.2\% to 52.5\%,
corresponding to an absolute gain of 23.3 percentage points over the
reproduced $\pi_{0.5}$ baseline. The improvement is consistent across the two
tasks: Bowl-Proximity increases from 25.0\% to 46.7\%, while
Eggplant-to-Pink-Plate increases from 33.3\% to 58.3\%. These results indicate
that the learned representation benefits both precision-sensitive
object-relative placement and longer manipulation sequences involving
semantic distractors.

The gains remain substantial under both spatial distribution changes.
Under Layout variation, GaussianDream++ doubles the pooled success rate from
25.0\% to 50.0\%, despite changes to the relative positions of targets and
distractors. Under Camera variation, success increases from 22.5\% to 42.5\%
when the global image projection changes while the underlying task remains
fixed. Together with the LIBERO-Plus results, these physical evaluations are
consistent with the intended role of Gaussian world supervision: World State
Tokens provide a more stable representation of scene structure, while World
Prediction Tokens encode interaction-relevant evolution that remains useful
throughout closed-loop execution.
\subsection{Qualitative Analysis}
\label{sec:qualitative}

\begin{figure*}[t]
    \centering
    \includegraphics[width=0.96\textwidth]{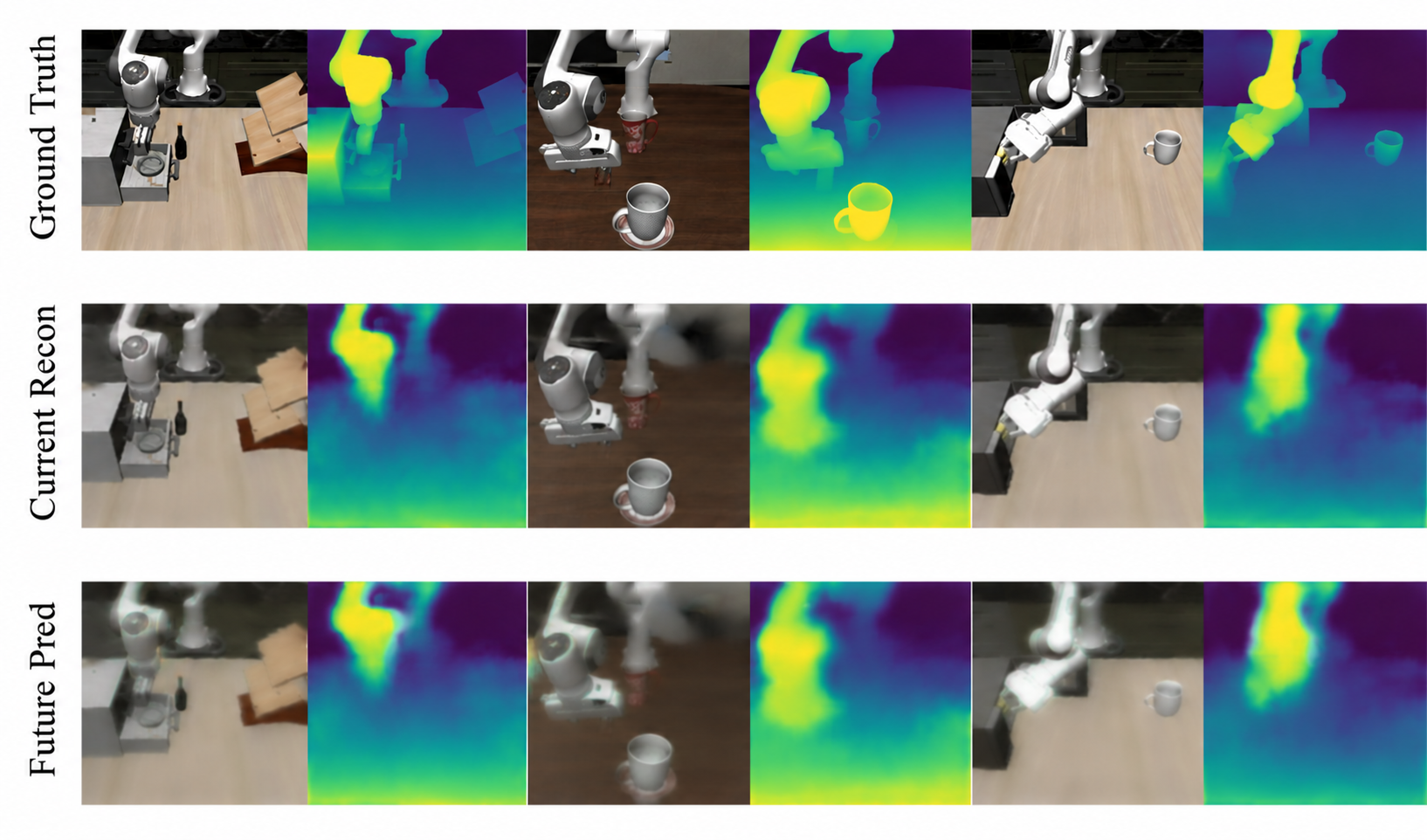}
    \caption{
    \textbf{Qualitative analysis of GaussianDream++.}
    We show Ground Truth, Current Reconstruction decoded from World State
    Tokens, and Future Prediction decoded jointly from World State Tokens and
    World Prediction Tokens. Current Reconstruction recovers the dominant
    robot configuration, workspace geometry, and object layout, whereas Future
    Prediction preserves the same coarse scene structure while representing
    short-horizon changes. The objective is not photorealistic future
    generation, but an action-relevant physical representation shaped by
    Gaussian world supervision.
    }
    \label{fig:vis}
\end{figure*}

We qualitatively examine whether GaussianDream++ encodes meaningful current
scene structure and short-horizon evolution. Figure~\ref{fig:vis} visualizes
RGB and depth decoded from the policy-native world representations across
three representative manipulation scenes. These outputs are produced only by
the training-time World Representation Head and are not consumed by the
Action Expert during deployment.

The Current Reconstructions preserve the dominant spatial organization of the
observations, including the robot configuration, tabletop structure, and the
approximate locations of task-relevant objects. Fine appearance details and
object boundaries remain imperfect, which is expected because the auxiliary
decoder is optimized to constrain the policy representation rather than to
serve as a standalone reconstruction system. The recovered depth structures
nevertheless show that the compact World State Tokens retain decodable
geometric information despite being substantially smaller than the dense
GaussianDream prefix.

Future Prediction is moderately less detailed than Current Reconstruction but
maintains consistent scene-level geometry and object correspondence. This
behavior is consistent with the coupled formulation: the Current World
provides the persistent Gaussian template, while World Prediction Tokens
encode residual short-horizon evolution instead of independently regenerating
the complete scene. The resulting predictions retain coarse physical changes
without requiring visually sharp future synthesis. These visualizations
provide an interpretability diagnostic of the learned representation; their
action relevance is evaluated quantitatively through the policy results and
component ablations in Sec.~\ref{sec:ablation}.

\subsection{Inference Efficiency}
\label{sec:efficiency}

GaussianDream++ uses rich Gaussian reconstruction and prediction only during
training. At deployment, the World Representation Head, Gaussian renderer,
future-target construction, and all auxiliary losses are removed; only the
World State Tokens and World Prediction Tokens remain in the native VLA
prefix. We measure end-to-end latency for one action-chunk prediction with
batch size one using the same local deployment pipeline for reproduced
$\pi_{0.5}$ and GaussianDream++.

\begin{table}[t]
    \centering
    \caption{
    \textbf{Inference-path and latency comparison.}
    Latency is reported per action chunk. The first block is measured using
    our matched local deployment pipeline. The GaussianDream value marked by
    $\dagger$ is reported by its original supplementary evaluation and is
    included only as a reference because the measurement environment may
    differ. Runtime Head indicates whether explicit Gaussian decoding or
    rendering is executed during deployment.
    }
    \label{tab:latency}
    \small
    \setlength{\tabcolsep}{5.5pt}
    \begin{tabular}{lcccc}
        \toprule
        Method
        & Runtime VGGT/TGE
        & World Tokens
        & Runtime Head
        & Latency \\
        \midrule

        \multicolumn{5}{c}{\textit{Matched local deployment}} \\
        \midrule

        $\pi_{0.5}$ reproduced
        & No
        & 0
        & No
        & 286 ms \\

        \textbf{GaussianDream++}
        & \textbf{No}
        & \textbf{20}
        & \textbf{No}
        & \textbf{330 ms} \\

        \midrule
        \multicolumn{5}{c}{\textit{Reported reference}} \\
        \midrule

        GaussianDream~\citep{zhang2026gaussiandream}$^\dagger$
        & Yes
        & 1024
        & No
        & 531 ms \\

        \bottomrule
    \end{tabular}
\end{table}

Under the matched local setting, GaussianDream++ increases action-chunk
latency from 286 ms to 330 ms, corresponding to 44 ms of additional
computation or approximately $1.15\times$ the latency of reproduced
$\pi_{0.5}$. This modest overhead arises from contextualizing 20 additional
World State and World Prediction Tokens inside PaliGemma; no Gaussian
primitive construction, rendering, or future rollout is executed online.

The architectural difference from GaussianDream lies in prefix construction.
GaussianDream removes its auxiliary Gaussian heads at deployment but still
uses VGGT and TGE to construct a unified 1024-token prefix from temporal
observations. GaussianDream++ instead forms its compact world representation
directly inside the VLA backbone and removes the dedicated runtime geometry
pathway. The reported GaussianDream latency provides a useful efficiency
reference, but an exact speedup claim requires all three policies to be
remeasured under identical hardware, precision, visual inputs, action horizon,
and warm-up settings.

\subsection{Ablation Studies}
\label{sec:ablation}

We conduct controlled ablations to answer two questions:
whether the gains of GaussianDream++ arise from additional token capacity or
from structured Gaussian world supervision, and which representation and
supervision components are responsible for the final performance.
All variants use the same PaliGemma/$\pi_{0.5}$ backbone, action objective,
training budget, and evaluation protocol; only the component specified by
each row is changed.

\begin{table*}[t]
    \centering
    \caption{
    \textbf{Ablation studies of GaussianDream++.}
    We report task success rate (\%) on the four LIBERO suites and the seven
    LIBERO-Plus distribution shifts. Structural ablations isolate token
    capacity, Current World supervision, Future Prediction, coupled decoding,
    and static consistency. Supervision ablations remove one Gaussian-world
    objective from the full model. LIBERO-Plus Overall follows the official
    task-level aggregation.
    }
    \label{tab:ablation}
    \scriptsize
    \setlength{\tabcolsep}{2.8pt}
    \resizebox{\textwidth}{!}{
    \begin{tabular}{lccccc|cccccccc}
        \toprule
        \multirow{2}{*}{Variant}
        & \multicolumn{5}{c|}{LIBERO}
        & \multicolumn{8}{c}{LIBERO-Plus} \\
        \cmidrule(lr){2-6}
        \cmidrule(lr){7-14}
        & Spatial
        & Object
        & Goal
        & Long
        & Avg.
        & Camera
        & Robot
        & Lang.
        & Light
        & BG
        & Noise
        & Layout
        & Overall \\
        \midrule

        $\pi_{0.5}$ reproduced
        & 98.8 & 98.2 & 98.0 & 92.4 & 96.9
        & 73.2 & 77.9 & 84.3 & 96.3
        & 95.5 & 89.9 & 87.7 & 85.5 \\

        \midrule
        \multicolumn{14}{c}{\textit{Structural ablations}} \\
        \midrule

        \makecell[l]{World Tokens w/o\\Gaussian Supervision}
        & 98.8 & 99.0 & 97.2 & 95.0 & 97.5
        & 75.0 & 73.5 & 85.0 & 97.5
        & 96.5 & 90.1 & 88.0 & 86.3 \\

        Current World only
        & 99.0 & 99.2 & 98.0 & 95.4 & 97.9
        & 77.0 & 73.2 & 86.2 & 98.6
        & 97.7 & 91.5 & 88.8 & 86.9 \\

        \makecell[l]{Current + Future,\\uncoupled}
        & 99.0 & 99.4 & 98.2 & 95.8 & 98.1
        & 78.0 & 73.0 & 86.8 & 98.8
        & 98.0 & 92.5 & 89.2 & 87.2 \\

        \makecell[l]{Coupled w/o\\Static Consistency}
        & 99.2 & 99.4 & 98.6 & 96.4 & 98.4
        & 79.2 & 72.9 & 87.3 & 99.0
        & 98.5 & 93.6 & 89.6 & 87.5 \\

        \midrule
        \multicolumn{14}{c}{\textit{Gaussian supervision ablations}} \\
        \midrule

        Full w/o RGB Rendering
        & 99.0 & 99.4 & 98.6 & 96.6 & 98.4
        & 79.1 & 72.8 & 87.1 & 98.5
        & 97.5 & 93.8 & 89.6 & 87.4 \\

        Full w/o Metric Depth
        & 98.8 & 99.2 & 98.0 & 96.0 & 98.0
        & 76.5 & 72.4 & 87.0 & 98.8
        & 98.2 & 91.8 & 87.7 & 86.9 \\

        Full w/o Alpha Supervision
        & 99.0 & 99.4 & 98.6 & 96.2 & 98.3
        & 79.0 & 72.8 & 87.2 & 98.9
        & 98.0 & 93.2 & 89.4 & 87.3 \\

        Full w/o Metric 3D Flow
        & 99.0 & 99.2 & 98.2 & 96.0 & 98.1
        & 78.4 & 72.7 & 87.1 & 98.9
        & 98.3 & 92.5 & 89.0 & 87.2 \\

        \midrule

        \rowcolor{gray!15}
        \textbf{Full GaussianDream++}
        & \textbf{99.2}
        & \textbf{99.6}
        & \textbf{99.0}
        & \textbf{96.6}
        & \textbf{98.6}
        & \textbf{80.1}
        & \textbf{73.0}
        & \textbf{87.4}
        & \textbf{99.1}
        & \textbf{98.8}
        & \textbf{94.2}
        & \textbf{90.0}
        & \textbf{87.8} \\

        \bottomrule
    \end{tabular}
    }
\end{table*}

\paragraph{Representation design.}
Adding World State Tokens and World Prediction Tokens without Gaussian
supervision improves LIBERO-Plus Overall from 85.5\% to 86.3\%, but remains
1.5 points below the full model. Thus, additional policy capacity alone does
not explain the gain. Constraining World State Tokens with Current World
reconstruction raises Overall to 86.9\%, with clear improvements under Camera
and Layout shifts. Introducing Future Prediction further improves Overall to
87.2\%, showing that short-horizon temporal supervision contributes beyond
current-scene grounding. Coupling Current World and Future Prediction over
shared Gaussian primitives increases performance to 87.5\%, while static
consistency yields the final 87.8\%. This progression supports the
role-separated design: World State Tokens provide the physical reference,
World Prediction Tokens encode short-horizon change, and static--dynamic
factorization suppresses unnecessary motion in persistent regions.

\paragraph{World-supervision objectives.}
Metric depth provides the strongest individual geometric constraint. Removing
it reduces Overall from 87.8\% to 86.9\%, with Camera and Layout decreasing
from 80.1\% and 90.0\% to 76.5\% and 87.7\%, respectively. Removing metric
3D flow reduces Overall to 87.2\% and particularly weakens Noise and Layout
robustness, indicating that rendering alone does not sufficiently constrain
interaction-induced motion. RGB rendering and alpha supervision are also
complementary: removing RGB primarily degrades Lighting and Background
performance, whereas removing alpha weakens visibility- and coverage-sensitive
shifts. The full objective performs best because appearance, metric geometry,
spatial coverage, and short-horizon motion constrain different aspects of the
same policy-native world representation.
\section{Conclusion}
\label{sec:conclusion}

We presented \textbf{GaussianDream++}, a compact, policy-native successor to
GaussianDream for physically grounded Vision-Language-Action control.
GaussianDream++ replaces the dense VGGT/TGE-based GaussianDream prefix with
20 World State Tokens and World Prediction Tokens embedded directly in the
PaliGemma backbone. During training, a lightweight World Representation Head
decodes these representations into a Current World and coupled Future
Prediction over shared Gaussian primitives. The role-separated design assigns
current-state grounding and short-horizon evolution to distinct token groups,
while static--dynamic factorization preserves persistent scene structure and
concentrates residual motion on interaction-relevant regions.

At deployment, the World Representation Head, Gaussian renderer, and all
auxiliary supervision branches are removed, and the Action Expert operates
directly on the compact world-token-augmented prefix. GaussianDream++ therefore
retains the dense geometric and temporal supervision introduced by
GaussianDream without its dedicated runtime VGGT/TGE pathway, online Gaussian
decoding, or future rollout. Under the matched GaussianDream-family protocol,
GaussianDream++ achieves 98.6\% on LIBERO and 87.8\% on LIBERO-Plus, with its
clearest gains under Camera and Layout shifts. It further improves pooled
real-robot success from 29.2\% to 52.5\% over reproduced $\pi_{0.5}$, while
incurring only modest latency over the base VLA policy. These results indicate
that Gaussian world supervision can be internalized into a substantially more
compact and structured policy representation, providing a practical route to
robust 3D-aware manipulation without a heavy deployed world model.

\clearpage
\bibliographystyle{bibstyle}
\bibliography{main}


\end{document}